\documentclass[conference]{IEEEtran}
\IEEEoverridecommandlockouts
\usepackage{cite}
\usepackage{amsmath,amssymb,amsfonts}
\usepackage{algorithmic}
\usepackage{graphicx}
\usepackage{textcomp}
\usepackage{xcolor}

\usepackage{multirow}
\usepackage{amssymb}
\usepackage{subfig}

\def\BibTeX{{\rm B\kern-.05em{\sc i\kern-.025em b}\kern-.08em
    T\kern-.1667em\lower.7ex\hbox{E}\kern-.125emX}}
\begin{document}

\title{FreConv: Frequency Branch-and-Integration Convolutional Networks
}

\author{\IEEEauthorblockN{1\textsuperscript{st} Zhaowen Li$^*$}
\IEEEauthorblockA{\textit{Institute of Automation} \\
\textit{Chinese Academy of Sciences$^1$,}\\
\textit{School of Artificial Intelligence} \\
\textit{University of Chinese Academy of Sciences$^2$}\\
Beijing, China \\
zhaowen.li@nlpr.ia.ac.cn}
\and
\IEEEauthorblockN{2\textsuperscript{nd} Xu Zhao}
\IEEEauthorblockA{\textit{Institute of Automation} \\
\textit{Chinese Academy of Sciences}\\
Beijing, China \\
xu.zhao@nlpr.ia.ac.cn}
\and
\IEEEauthorblockN{3\textsuperscript{rd} Peigeng Ding$^*$\thanks{$^*$These authors contributed to the work equllly and should be regarded as co-first authors.}}
\IEEEauthorblockA{\textit{Institute of Automation} \\
\textit{Chinese Academy of Sciences$^1$,}\\
\textit{School of Artificial Intelligence} \\
\textit{University of Chinese Academy of Sciences$^2$}\\
Beijing, China \\
peigeng.ding@nlpr.ia.ac.cn}
\and
\IEEEauthorblockN{4\textsuperscript{th} Zongxing Gao}
\IEEEauthorblockA{\textit{School of Art and Design} \\
\textit{Beijing Institute of Graphic Communication}\\
Beijing, China \\
Gaoxinxino@163.com}
\and
\IEEEauthorblockN{5\textsuperscript{th} Yuting Yang}
\IEEEauthorblockA{\textit{Institute of Automation} \\
\textit{Chinese Academy of Sciences}\\
Beijing, China \\
yangyuting02@baidu.com}
\and
\IEEEauthorblockN{6\textsuperscript{th} Ming Tang}
\IEEEauthorblockA{\textit{Institute of Automation} \\
\textit{Chinese Academy of Sciences}\\
Beijing, China \\
tangm@nlpr.ia.ac.cn}
\and
\IEEEauthorblockN{7\textsuperscript{th} Jinqiao Wang}
\IEEEauthorblockA{\textit{Institute of Automation} \\
\textit{Chinese Academy of Sciences$^1$,}\\
\textit{Peng Cheng Laboratory$^2$} \\
Beijing, China \\
jqwang@nlpr.ia.ac.cn}
}

\maketitle

\begin{abstract}
Recent researches indicate that utilizing the frequency information of input data can enhance the performance of networks. However, the existing popular convolutional structure is not designed specifically for utilizing the frequency information contained in datasets. In this paper, we propose a novel and effective module, named FreConv (frequency branch-and-integration convolution), to replace the vanilla convolution. FreConv adopts a dual-branch architecture to extract and integrate high- and low-frequency information. In the high-frequency branch, a derivative-filter-like architecture is designed to extract the high-frequency information while a light extractor is employed in the low-frequency branch because the low-frequency information is usually redundant. FreConv is able to exploit the frequency information of input data in a more reasonable way to enhance feature representation ability and reduce the memory and computational cost significantly.  Without any bells and whistles, experimental results on various tasks demonstrate that FreConv-equipped networks consistently outperform state-of-the-art baselines.
\end{abstract}

\begin{IEEEkeywords}
Convolutional structure, image classification, object detection, instance segmentation
\end{IEEEkeywords}

\section{Introduction}
Deep learning has shown excellent performance on various computer vision tasks~\cite{2015ImageNet,he2016deep,ren2017faster,chen2022obj2seq,chen4299967devil}. Recently, some researchers \cite{li2022transfering,chen2019drop,zhang2019making,li2020wavelet} have indicated that utilizing the frequency information of input data can improve the performance of networks. The research community has devoted significant efforts to introducing frequency information into CNNs to acquire better fitting. For exploiting frequency information, the prior work \cite{li2022transfering,zhang2019making,2020Delving,li2020wavelet} integrated the classic low-pass filter with the downing-sampling in CNNs to obtain low-frequency features. These methods demonstrated the performance superior to the baselines' \cite{he2016deep} because their replacement of pooling layers conforms to the Nyquist Theorem \cite{nyquist1928certain}. Nevertheless, their networks focus only on low-frequency information and are not able to exploit the information of \emph{whole} frequencies contained in datasets. If high-frequency information is crucial for some tasks, these methods may not be effective. Morover, OctConv \cite{chen2019drop} argued that the output features of a convolutional layer can be decomposed into features of different frequencies and took into account the high-frequency information. To extract multi-frequency features, OctConv designed an architecture to store and process high- and low-frequency features. However, OctConv did not specifically design convolutional structure for utilizing the frequency information contained in datasets.

\begin{figure}[t]
    \small
    \centering
    
    \includegraphics[width=3in]{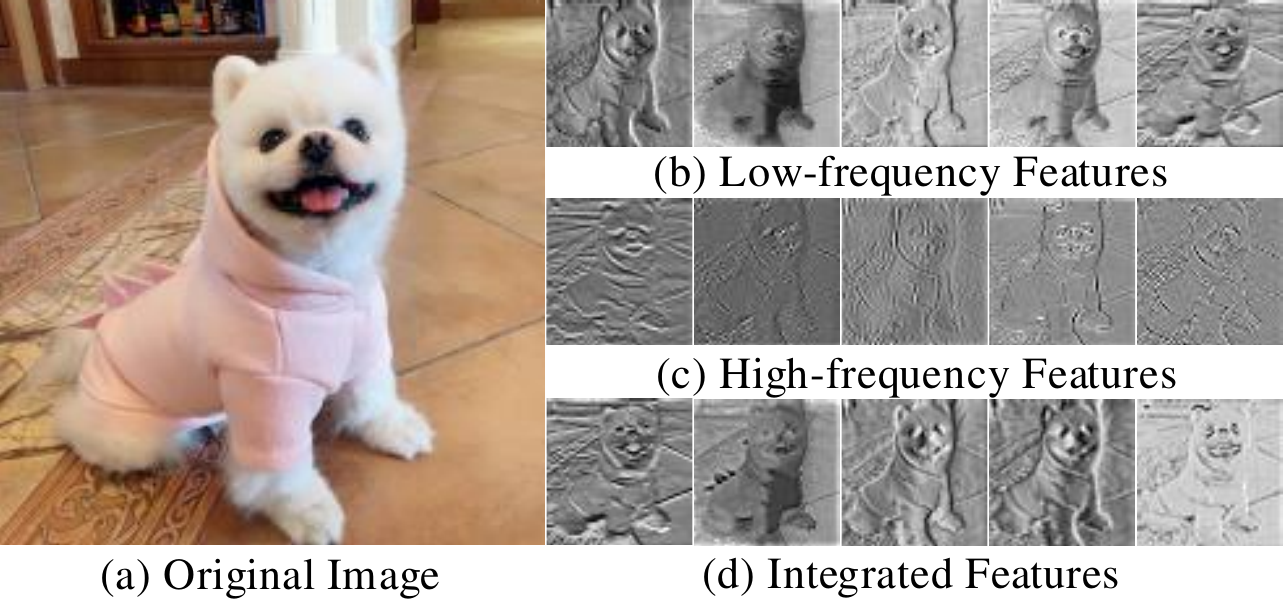}
    \includegraphics[width=2.9in]{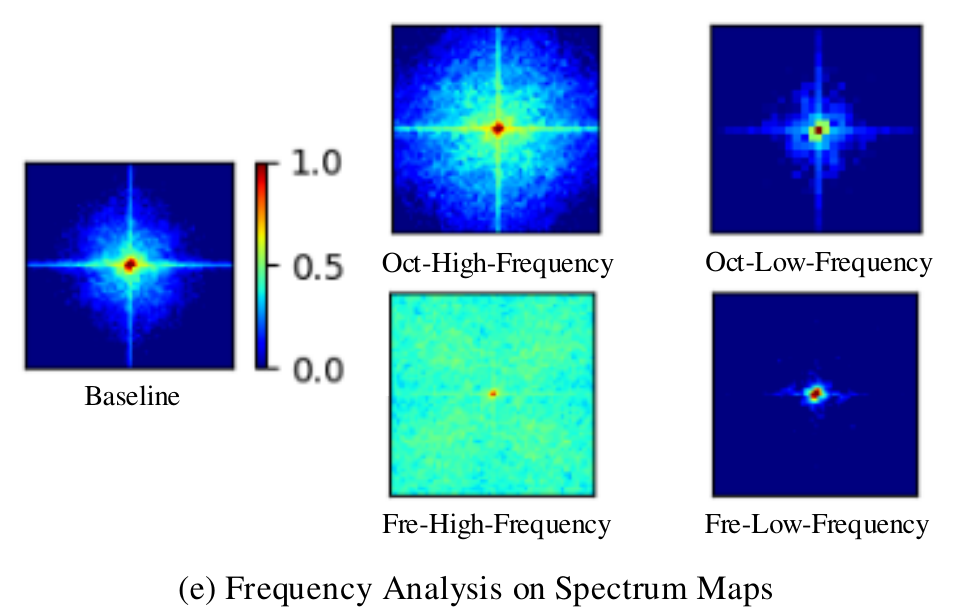}
    \vspace{-1mm}
    \caption{   (a) Original input image.   (b) Low-frequency output feature maps. (c) High-frequency output feature maps. (d) Integrated output feature maps. (e) Frequency analysis on spectrum maps. 1000 images are sampled randomly from ImageNet~\cite{2015ImageNet} and input to OctConv \cite{chen2019drop} and our FreConv-ResNet50, respectively. The spectrum maps of features of 1000 images from the first $3\times3$ convolutional layer of vanilla ResNet50 \cite{he2016deep}, OctConv-ResNet50, and our FreConv-ResNet50, respectively, are averaged and shown. ``Baseline'' refers to the first $3\times3$ convolutional layer of vanilla ResNet50~\cite{he2016deep}. See the text for details. } 
    \label{highlow}
    \vspace{-4mm}
\end{figure}

In order to overcome the above drawback, in this paper, we propose the frequency branch-and-integration convolution module (FreConv) to extract and integrate high- and low-frequency features effectively and efficiently to exploit the \emph{whole} frequency information. FreConv is able to adaptively exploit frequency information of input data in a more reasonable way, enhancing the feature representation ability notably with significantly reducing of computational and memory costs. FreConv mainly consists of three modules, feature split, feature extraction, and feature integration. In the feature split module, a dual-branch attention module is designed to adaptively adjust the activation value of input features to make the activation features suitable for dual-branch high- and low-frequency feature extraction and fuse the inter-channel information. To extract features of different frequencies, we utilize a dual-branch architecture to extract high- and low-frequency features through learning in the feature extraction component. In the high-frequency branch of this module, a derivative-filter-like architecture is designed to extract the high-frequency information. In essential, such architecture will produce a series of derivative  filters after learning, extracting the information of various high-frequencies. In addition, it is known that the low-frequency information describes a smoothly changing structure \cite{devalois1990spatial} and carries less information according to Shannon Information Theory \cite{Shannon1948IEEE,Perlis1965Principles}. Hence, in the low-frequency branch of feature extraction, the point-wise convolution \cite{szegedy2015going} is adopted to avoid redundant parameters and calculations. In the feature integration module, we integrate the extracted frequency features with the point-wise summation to avoid losing crucial information. 

To illustrate the effect of our FreConv on the extraction of high- and low-frequency information, we replace all $3\times3$ convolutions of ResNet50 by FreConvs which is denoted by FreConv-ResNet50 in Fig.~\ref{highlow}, and visualize the input image and some output high- and low-frequency features of the first FreConv. The results are shown in Fig.~\ref{highlow}. It is seen from Fig.~\ref{highlow}(b)(c) that our FreConv extracts the high- and low-frequency information of the input image, and Fig.~\ref{highlow}(d) is regarded as the feature with enhanced representation ability (see Sec.~\ref{experiments}). To further demonstrate the effectiveness of our FreConv, we adopt the Fourier transform to analyze the average energy spectrum map of the 30k features. It is seen from Fig.~\ref{highlow}(e) that our FreConv is able to extract the low-frequency information better and the information of much richer high-frequencies than OctConv does.

To further validate the universality of our FreConv, we further integrate FreConv into representative backbones, ResNet \cite{he2016deep}, VGG \cite{simonyan2014very}, and DenseNet \cite{huang2017densely}, and test their performance. It is noted that our FreConv is a plug-and-play replacement for the vanilla convolution. Experimental results show that the backbones plus our module are superior to the state-of-the-art methods at lower memory and computational costs. In summary, our main contributions are as follows:
\begin{itemize}
	\item We design a plug-and-play module, FreConv, that exploits the frequency information of input data in a more reasonable way with significant computational and memory cost reduction for various tasks.
	\item We propose a dual-branch architecture to extract the features of various frequencies, and integrate these features by an efficient and effective approach.
	\item We integrate FreConv into various CNN backbones for diverse tasks, and all integrations outperform the state-of-the-art methods or baselines.
\end{itemize}

\section{Approach}
\label{approach}
The pipeline of our proposed FreConv is shown in Fig.~\ref{fig:FSConv}, which mainly consists of three components: feature split, feature extraction, and feature integration. In this section, we first introduce the feature extraction module, and then explain the rest of the modules. Finally, we analyze how to use this module to build networks in detail.

\subsection{High- and Low-frequency Feature Extraction}
\label{fp}

The high-frequency information describes the rapidly changing fine details \cite{devalois1990spatial} and carries more quantities of information \cite{Shannon1948IEEE}. The complex information is various and vanilla learned convolution has difficulty autonomously extracting the information without specific constraints or structures. Hence, we propose a derivative-filter-like architecture, which consists of different convolutions to extract high-frequency features $X_{high}$ through learning in the high-frequency extraction (HFE) branch. Moreover, Ma \emph{et al.} \cite{ma1998derivative} proved that the Difference of Exponential is a derivative filter, which has the attributes of the high-pass filter. According to \cite{ma1998derivative}, the derivative filter obtained by the equation is shown in Eq (\ref{huv}), where $ 0 \leq u \leq H$, $0 \leq v \leq W$, and $\sigma$ is the scale factor, which is related to the size of the convolution kernel. $\alpha(\cdot)$ represents the weighting coefficients of exponential filters, and they can be given with Eq (\ref{sig}). From the Eq (\ref{huv}), we argue that exponential filters with different sizes construct different derivative filters, which will affect the performance of the network.
\begin{equation}
    H(u,v) = \alpha(\sigma_0)e^{-\frac{\sqrt{u^2+v^2}}{\sigma_0^2}} - \alpha(\sigma_1)e^{-\frac{\sqrt{u^2+v^2}}{\sigma_1^2}}  \label{huv}
\end{equation}
\begin{equation}
    \alpha(\sigma) = \frac{1+e^{-\frac{1}{\sigma}}}{1-e^{-\frac{1}{\sigma}}} \label{sig}
\end{equation}

\begin{figure}
\begin{center}
\includegraphics[width=1\linewidth]{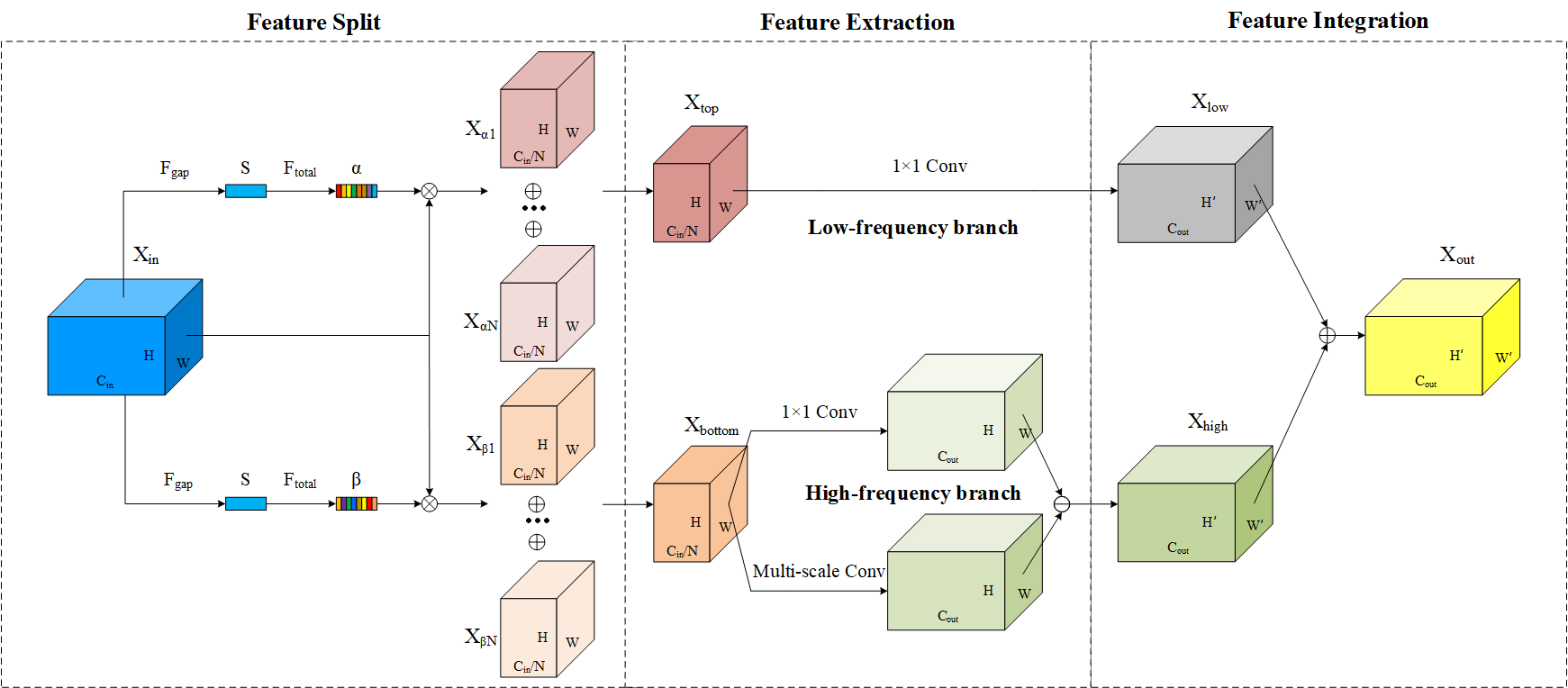} 
\end{center}
\vspace{-3mm}
\caption{The overview of our FreConv. The module has two branches: the above branch extracts low-frequency information, and the below branch extracts high-frequency information. Finally, the extracted features from the two branches are integrated in the end.  }
\label{fig:FSConv}
\vspace{-2mm}
\end{figure}

To make the network adaptively extract high-frequency information that is useful for tasks, the multi-scale convolution is leveraged as the first exponential filter while fixing the point-wise convolution as the second in the right of Eq (\ref{huv}). As shown in Fig.~\ref{fig:FSConv}, FreConv makes the difference between the multi-scale convolution and point-wise convolution construct the derivative-filter-like architecture according to Eq (\ref{huv}). \emph{The derivative-filter-like architecture is initialized with the parameters of the exponential filter}. 

Here, we design Multi-scale Conv (More details can refer to Appendix), which combines the different sizes of convolution to extract information. Input features $X_{bottom}$ are completely passed through each convolution and output features. Then, these features are concatenated by channel. The Multi-scale Conv is initialized with the parameters of the exponential filter and combined with the point-wise convolution to construct the derivative-filter-like architecture. This architecture can produce a series of derivative  filters through learning. The architecture essentially differs from Inception~\cite{szegedy2015going}.

Meanwhile, the high-pass filter passes the high-frequency information while suppressing the low-frequency information. The low-frequency information needed by the task has to pass through the low-frequency extraction (LFE) branch. It is known that the low-frequency information describes the smoothly changing structure \cite{devalois1990spatial} and carries fewer quantities of information \cite{Shannon1948IEEE}. OctConv reduces the redundancy of low-frequency information by reducing resolution. Different from OctConv, we apply the point-wise convolution operation to extract low-frequency information in order to reduce redundancy and obtain low-frequency features $X_{low}$ in the LFE branch. In order to employ the larger convolution (compared with $3\times3$ convolution) without increasing parameters and FLOPs in the HFE branch, we adopt the dilated or group convolution and propose two options: dilated convolution kernel (DCK), and group convolution kernel(GCK).

\noindent\textbf{Dilated Convolution Kernel}. The dilated convolution can make the $3\times3$ convolution acquire a larger receptive field without changing the number of parameters and FLOPs. Hence, we can obtain the receptive field of $K \times K$ convolution kernels corresponding to $3\times3$ convolution kernels with a dilated rate $r \in \{1, 2, 3, 4\}$  as Eq (\ref{diaqua}), where $K \in \{3, 5, 7, 9\}$ is the size of convolution kernels. Unlike \cite{2018DeepLab,chen2018encoder,2020Scale}, where the dilated convolution is only added to the end of the network structure, and we adopt the dilated convolution from the beginning of the network.
\begin{equation}
    K = 3 + 2(r - 1) \label{diaqua}
\end{equation}
\noindent\textbf{Group Convolution Kernel}. The group convolution can make the number of parameters and calculations of $K \times K$ convolution be basically the same as that of $3\times3$ vanilla convolution. We can utilize the larger convolution kernel according to Eq (\ref{eq9}). From this equation, we can conveniently employ a larger convolution kernel by setting the group as $g_2$, where $g_1, g_2 \in \{2, 4, 8, 16\}$.
\begin{equation}
    g_2 \approx \frac{K_{2}^{2} \times g_1}{K_{1}^{2}}   
    \label{eq9}
\end{equation}

\subsection{Attention-based Feature Split}
\label{sec31}

To make the input activation features suitable for the dual-branch high- and low-frequency feature extraction and fuse the inter-channel information.
In this paper, we propose a dual-branch attention module, named the feature split module, to dynamically adjust the activation value of input features to make the input activation features suitable for dual-branch high- and low-frequency feature extraction. Given an input feature map: $X_{in} \in R^{ C \times H \times W}$, we use global average pooling to generate channel-wise statistics $S \in R^{C\times 1\times1}$ as shown in Eq (\ref{eq1}).
\begin{equation}
    S = F_{gap}\left(X_{in}\right)=\frac{1}{H \times W}\sum_{i=1}^H \sum_{j=1}^W X_{in}\left(i,j\right) \label{eq1}
\end{equation}

Meanwhile, we adopt a dual-branch method to make a single branch into two branches output. Where $\alpha$, $\beta \in R^{C \times1\times1}$ respectively represent the weights of the two branches, and the whole process of acquiring them are expressed with Eq (\ref{eqq2}), (\ref{eq2}), $\eta$  refers to the sigmoid function and $\delta$ is the ReLU activation. $F_{conv1\_1}, F_{conv1\_2}, F_{conv2\_1}$, and $F_{conv2\_2}$ is $1\times1$ convolution operations.
\begin{equation}
    \alpha=F_{total_{1}} \left( S\right)=\eta \left(F_{conv1\_2}\left(\delta \left(F_{conv1\_1}\left(S\right)\right)\right)\right) \label{eqq2} 
\end{equation}
\begin{equation}
    \beta=F_{total_{2}} \left( S\right)=\eta \left(F_{conv2\_2}\left(\delta \left(F_{conv2\_1}\left(S\right)\right)\right)\right)
        \label{eq2}
\end{equation}

We multiply the input feature map by $\alpha$ and $\beta$ to get the weighted features $X_{\alpha}$ and $X_{\beta}$ , and they can be given with Eq (\ref{eqq3}), (\ref{eq3}). 
\begin{equation}
    X_{\alpha}=\alpha * X  \label{eqq3}
\end{equation}
\begin{equation}
    X_{\beta}=\beta * X  \label{eq3}
\end{equation}

As the Fig.~\ref{fig:FSConv} shows, the top branch is treated as the low-frequency branch, and the bottom branch is the high-frequency branch. To take into account the correlation between feature channels, we split each branch into two parts by channel by default. Note that, this way of splitting is scalable. Here we divide the feature channel into $N$ parts, and $N \in \{2,4,8,16\}$, each part of the same size, and get $X_{\alpha1}, ..., X_{\alpha N}, X_{\beta1}, ..., X_{\beta N}\in R^{C/N \times H \times W}$. Then we add these features along the channel to achieve the purpose of fusing inter-channel information. Finally we obtain the split features $X_{top}, X_{bottom} \in R^{C/N \times H \times W}$. The whole process can be expressed with Eq (\ref{eqq4}), (\ref{eq4}).
\begin{equation}
    X_{top}= X_{\alpha1}+...+X_{\alpha N} \label{eqq4}
\end{equation}
\begin{equation}
   X_{bottom}=X_{\beta1}+...+X_{\beta N} \label{eq4}
\end{equation}

The attention mechanism can strengthen the features that are suitable for extracting frequency information while weakening the inappropriate features. Notably, the module is complementary to the SE module \cite{hu2018squeeze} and we prove this in the Experiments (see Table \ref{tab:my_label}).

\subsection{Feature Integration}
\label{sec3.3}
After we obtained the extracted high- and low-frequency information through learning in Sec.~\ref{fp}, we combine this information to express the input information for the next stage.  Therefore, we integrate the extracted frequency features with the point-wise summation to avoid losing crucial information and enhance the feature representation ability. Finally, we get output features $X_{out}$ as Fig.~\ref{fig:FSConv}.

\subsection{FreConv-equipped Networks}
The network employs a dual-branch to extract frequency features, which requires the high integrity of the input frequency information. However, max-pooling can easily lose information \cite{zhang2019making}. Hence, without changing the basic structure of the network, we replace the max-pooling operation with the strided-convolution to ensure the relative integrity of the information. In the first stage, the maximum size of convolution kernel used is $9 \times9$. However, the resolution of features decreases as the network deepens, and our FreConv-equipped ResNet (FreConv-ResNet) network removes the previous largest convolution in the current stage compared to the previous stage. This design can be extended to other network models, such as VGG, and DenseNet.

\section{Experiments}
\label{experiments}
\subsection{Experimental Setups}
\label{setup}

\textbf{Image Classification}. For image classification, we use the most popular dataset ImageNet \cite{2015ImageNet} for all experiments. The experimental settings are fully following  \cite{he2016deep}. 

\noindent \textbf{Object Detection}. We train our model on the COCO dataset. For object detection, the setting of experiments strictly follows Faster R-CNN~\cite{ren2017faster}.  

\noindent \textbf{Instance Segmentation}. Mask R-CNN \cite{he2020mask} is adopted as the segmentation framework on COCO.

\begin{table}
\small
\renewcommand\arraystretch{0.9}
	\setlength{\tabcolsep}{0.2pt}
	\caption{Validation accuracy rate, parameters, and calculations comparison results of FreConv on ImageNet with other SOTAs. } \label{tab:my_label} 
    \scalebox{0.86}{
        \begin{tabular}{llcccccc}
        \hline
            Backbone&Method&Params&Reduced&FLOPs&Reduced&Top-1\\
        \hline  
            \multirow{13}{*}{ResNet50}&ResNet-Baseline \cite{he2016deep} &25.56M&\textbf{-}& 4.14G&\textbf{-}&75.89 \\
        &SE-ResNet \cite{hu2018squeeze} & 28.07M&+9.82&4.15G&+0.24&76.82 \\
        &SE-ResNeXt \cite{hu2018squeeze,xie2017aggregated}  & 27.56M&+7.82&4.32G&+4.35&78.05 \\
        &Anti-aliased-ResNet \cite{zhang2019making}&25.56M&0.00&5.16G&+24.64&76.84\\
        &WResNet-Haar \cite{li2020wavelet}&25.56M&0.00&5.16G&+24.64&76.89\\
        &Adaptive-anti-ResNet \cite{2020Delving}&25.56M&0.00&5.17G&+24.88&77.01\\
        &OctConv-ResNet-$\alpha\frac{1}{2}$ \cite{chen2019drop}&25.56M&0.00& 2.40G&\textbf{-}42.00&76.80 \\
        &ResNeXt50 \cite{xie2017aggregated}&25.03M&-2.07&4.30G&+3.87&77.22\\
        &PyConvResNet \cite{duta2020pyramidal}&24.85M&-2.78&3.88G&-6.28&77.44\\
        &\textbf{Ours-GCK-ResNet-N-4}&\textbf{16.56M}&\textbf{-35.21}&\textbf{2.69G}&\textbf{-35.02}&\textbf{77.13}\\
        & \textbf{Ours-DCK-ResNet50-N-2}&\textbf{18.50M}& \textbf{-27.62}&\textbf{2.98G}&\textbf{-28.02} &\textbf{77.53} \\
        &\textbf{Ours-GCK-ResNet-N-2}&\textbf{18.71M}&\textbf{-26.80}&\textbf{3.07G}&\textbf{-25.85}&\textbf{77.80}\\
        &\textbf{SE-Ours-GCK-ResNet-N-2}&\textbf{21.22M}&\textbf{-16.98}&\textbf{3.08G}&\textbf{-25.60}&\textbf{78.80}\\
        \hline
        \multirow{2}{*}{ResNet101}&ResNet-Baseline \cite{he2016deep}&44.55M&\textbf{-}&7.85G&\textbf{-}&77.13\\
        &\textbf{Ours-GCK-ResNet-N-2}&\textbf{31.94M}&\textbf{-28.31}&\textbf{5.64G}&\textbf{-28.15}&\textbf{78.75}\\
        \hline
        \multirow{2}{*}{ResNet152}&ResNet-Baseline \cite{he2016deep}&60.19M&\textbf{-}&11.58G&\textbf{-}&77.90\\
        &\textbf{Ours-GCK-ResNet-N-2}&\textbf{42.85M}&\textbf{-28.81}&\textbf{8.22G}&\textbf{-29.02}&\textbf{79.27}\\
        \hline
        \multirow{2}{*}{DenseNet}&DenseNet121-Baseline \cite{huang2017densely} &7.98M&\textbf{-}& 2.88G&\textbf{-}&74.98 \\
        &\textbf{Ours-GCK-DenseNet-N-2}&\textbf{6.89M}&\textbf{-13.66}&\textbf{2.33G}&\textbf{-19.10}&\textbf{75.81}\\
        \hline
        \multirow{2}{*}{VGG}&VGG16-Baseline \cite{simonyan2014very} & 138.37M&\textbf{-}&15.53G&\textbf{-}&72.98 \\
        &\textbf{Ours-GCK-VGG-N-2}&\textbf{129.59M}&\textbf{-6.35}&\textbf{7.67G}&\textbf{-50.61}&\textbf{74.12}\\
        \hline
        \end{tabular}}
      \vspace{-0.8cm}
\end{table}

\subsection{Comparing with SOTAs on ImageNet}
To show the effectiveness of the proposed FreConv, in this section, we replace the widely-used vanilla $3\times3$ convolution with the proposed FreConv. The upgraded networks have only one global hyper-parameter $N$. Many methods use different and complex training settings, for fair comparisons, we re-implement these methods without tricks.

In Table \ref{tab:my_label}, it can be observed that most methods that utilize frequency information have significantly increased calculations. Moreover, none of these methods decrease the parameters. The performance of FreConv-ResNet50 is better than the baseline by 1.91\% in terms of top-1 accuracy with parameters and FLOPs reduced by 26.80\% and 25.85\%, when we adopt the GCK method and set $N$ to 2. We compare FreConv-ResNet with a set of state-of-the-art methods: OctConv-ResNet50 \cite{chen2019drop}, anti-aliased-ResNet50 \cite{zhang2019making}, WaveCNet \cite{li2020wavelet}, adaptive-anti-aliased-ResNet50 \cite{2020Delving}, SE-ResNet \cite{hu2018squeeze}, SE-ResNeXt \cite{xie2017aggregated}, PyConvResNet \cite{duta2020pyramidal}. Experimental results show that the backbones plus our module are superior to the state of the art at lower memory and computational cost. Meanwhile, FreConv is designed to replace the vanilla convolution. The existing backbones often adopt SE module \cite{hu2018squeeze} to enhance the performance of networks. Our FreConv is orthogonal to the SE module, the combination of FreConv and the SE module can further enhance the performance of the network (from 77.80 to 78.80), and the result outperforms the combination of ResNeXt and the SE module by 0.7 (from 78.8 to 78.1) with nearly 30\% FLOPs and 30\% parameters dropped. We also migrate our method to other models, such as DenseNet and VGG. Despite reducing the amount of FLOP by 50.61\%, our method shows superior performance to the VGG baseline (from 72.98 to 74.12). DenseNet-121 contains only 7.98M parameters, about 5.77\% of VGG16, while it performs 2\% more accurately. Although it would be more challenging to improve performance, FreConv-DenseNet121 still outperforms the baseline (from 74.98 to 75.81).

\subsection{Object Detection}
We further evaluate our FreConv's generalizability on the object detection task of the COCO dataset. The two-stage Faster R-CNN is used as our framework. As shown in Table \ref{tab:object_detection_size}, FreConv-GCK-ResNet50-N-2 outperforms its counterparts by 2.0\% on mean average precision ($mAP$) and 1.5\% on mean average recall ($mAR$) and has a large margin of improvement over its counterparts by 1.6\%, 2.0\%, and 2.6\% on $AP$ for small, medium, and large objects, respectively. The improvement of $AR$ for small, medium, and large objects are 1.9\%, 1.2\%, and 2.1\%, respectively. These results confirm the effectiveness of the proposed FreConv method.

\begin{table}[t]
\centering
\renewcommand{\arraystretch}{0.25}
\caption{Average Precision ($AP$) and Average Recall ($AR$)
    of object detection with different sizes on the MS COCO dataset.} \vspace{-0.1cm}
\resizebox{\columnwidth}{!}{
  \begin{tabular}{lccccc} \hline
                   &     &\multicolumn{4}{c}{Object size }\\ \cline{3-6}
                   &    & Small & Medium & Large & All  \\  \hline
    ResNet50 & \multirow{3}{*}{$AP (\%)$} & 21.1 & 39.2 & 45.8 & 35.8 \\
    FreConv-ResNet50 &            & \textbf{22.7}  & \textbf{41.2}  & \textbf{48.4}  & \textbf{37.8}  \\
    Improve.  &            & +1.6 & +2.0 & +2.6 & +2.0 \\ \hline
    ResNet50 & \multirow{3}{*}{$AR (\%)$} & 31.7 & 54.1 & 62.1 & 50.0 \\
    FreConv-ResNet50  &           & \textbf{33.6} & \textbf{55.3} & \textbf{64.2} & \textbf{51.5} \\
    Improve.  &            & +1.9 & +1.2 & +2.1 & +1.5 \\ \hline
  \end{tabular}
}\vspace{-0.1cm}

\label{tab:object_detection_size}
\end{table}

\begin{table}[t]
\centering
\renewcommand{\arraystretch}{0.3}
\caption{Performance of instance segmentation on MS COCO validation set using FreConv with different scales.} \vspace{-0.1cm}
\resizebox{\columnwidth}{!}{
  \begin{tabular}{lcccccc}\hline
   Backbone              & $AP$&$AP_{50}$&$AP_{75}$&$AP_{S}$&$AP_{M}$&$AP_{L}$  \\ \hline
   ResNet50                & 33.7  &54.4&35.6&17.5&36.7&45.7          \\ \hline
   FreConv-ResNet50    & \textbf{35.5} &\textbf{57.6} &\textbf{37.8} &\textbf{20.2} &\textbf{39.0} &\textbf{47.2}             \\
   \hline
  \end{tabular}
} \vspace{-0.4cm}

\label{segmentation}
\end{table}

\begin{table}[t]
\small
\renewcommand\arraystretch{0.85}
	\begin{center}
	\setlength{\tabcolsep}{0.5pt}
	\vspace{-0.2cm}
	\caption{Results of the ImageNet classification experiments. Starting from our baseline, we gradually add feature split, down-sampling, feature integration, and frequency feature extraction scheme in our FreConv-ResNet for ablation studies. Where  ``$3\times3$'' operation refers to the combination of $3\times3$ convolution and point-wise convolution,``Same" operation refers to the LFE and HFE branch adopting the same operation. LFE: low-frequency extraction, HFE: high-frequency extraction. We report the Top-1 accuracy (\%).}\label{tb:exp_ablation} 
	\scalebox{0.82}{
	\begin{tabular}{c|cc|cc|cc|ccc|cc|c}
		\hline
		 \multirow{2}{*}{Method} &\multicolumn{2}{|c|}{Split}  & \multicolumn{2}{|c|}{Sampling} & \multicolumn{2}{|c|}{LFE} &  \multicolumn{3}{|c|}{HFE}  & \multicolumn{2}{|c|}{Integration} &\multirow{2}{*}{Top-1}   	\\
		\cline{2-12}
		&Direct&Our&Pool&Strided&1 $\times$ 1&Same&3 $\times$3 &Same&Multi&Sum&Attention&\\
		\cline{1-13}
		 (a)  & - & - & - 					&  	-&-	& -	& -& -& -& -& -&75.89\\
		\cline{1-13}
				 (b)  	 & $\checkmark$ &	& $\checkmark$						&		& $\checkmark$	& 	& $\checkmark$&	&& $\checkmark$ & 	&75.82 \\
		 (c) 	& $\checkmark$ &		& $\checkmark$   &	    & $\checkmark$  & 	&$\checkmark$ 	& 	& &	& $\checkmark$   &76.12\\
		 (d)  &  & $\checkmark$	& $\checkmark$						&		& $\checkmark$	& 	& $\checkmark$	&&& $\checkmark$ & 	&76.91 \\
		 (e) 	&  & $\checkmark$	& $\checkmark$						&		& $\checkmark$	& 	& $\checkmark$	&&&  &$\checkmark$ 	&76.28 \\
		 (f) 	&  & $\checkmark$	&					& $\checkmark$			& $\checkmark$	& 	& $\checkmark$	&&& $\checkmark$  & 	&77.23 \\
		 (g) 	&  & $\checkmark$	&					& $\checkmark$			& 	&$\checkmark$ 	&& $\checkmark$	&& $\checkmark$  & 	&77.19 \\
		 (h) 	&  & $\checkmark$	&					& $\checkmark$			& $\checkmark$ 	& 	& &	&$\checkmark$ & $\checkmark$  & 	&\textbf{77.80} \\
		\hline	
						
	\end{tabular}
	}
	\end{center}
    \vspace{-0.8cm}

\end{table}

\subsection{Instance Segmentation}
Mask R-CNN \cite{he2020mask} is adopted as the instance segmentation method, and the performance of instance segmentation on the COCO dataset is shown in  Table \ref{segmentation}. The FreConv-GCK-ResNet50-N-2 based method outperforms its counterparts by 1.8\% on mask $AP$ and 3.2\% on mask $AP_{50}$. The performance gains on objects with different sizes are also demonstrated. The improvement of $AP$ for small, medium, and large objects are 2.7\%, 2.3\%, and 1.5\%, respectively.

\subsection{Ablation Study}

In this section, we conduct an ablation study to investigate the relative effectiveness of each part in FreConv. The baseline methods (Table \ref{tb:exp_ablation}(a)) are conducted with ResNet50 on the ImageNet dataset. The experimental results are shown in Table \ref{tb:exp_ablation}. In Table \ref{tb:exp_ablation}, we use GCK methods to reduce the computation complexity in all listed experiments and we set $N$ to 2 in the feature split module.

\noindent \textbf{Feature Extraction Methods}: In this part, the down-sampling module, the feature split and feature integration method adopt the same strategy. The network topology and parameters of the LFE branch are consistent with those of the HFE branch using $3\times3$ convolution grouped 2 in method (g). Meanwhile, the method (f) adopts $1\times1$ convolution in the LFE branch and employs the $3\times3$ convolution grouped 2 adding point-wise convolution in the HFE branch. However, method (f) outperforms method (g) (from 77.19 to 77.23) although method (g) has more parameters and calculations than method (f) because the low-frequency branch obtains more parameters, we argue that allocating more parameters in the low-frequency branch would lead to over-fitting. The performance of method (f) is much lower than that of method (h) (from 77.23 to 77.80). The result shows that $3\times3$ learned convolution is insufficient in the extraction of various high-frequency information, and the derivative-filter-like architecture can help the network to better extract the high-frequency information useful for tasks. Moreover, we try to fix the initialization parameters of the derivative-filter-like architecture so that these parameters do not participate in network training. However, although the fixed-parameter network can work, the performance of the learned-parameter network is much better than it. The derivative-filter-like architecture that can be adjusted through learning is better than the fixed coefficients.

\noindent \textbf{Feature Split and Feature Integration Methods}: In this part, the down-sampling module and the frequency extraction branch use the same strategy. For methods (c) and (e), we adopt popular parameter-free soft-attention feature integration module according to Transformer \cite{vaswani2017attention}. In (b), we fuse the features by direct point-wise summation which is mentioned in Section \ref{sec3.3}. We find that the attention module outperforms point-wise summation operation (c) while the feature split module of methods (b) and (c) adopts the directly split feature operation (from 76.12 to 75.82, the difference is 0.3). In contrast, the performance of this module (e) is lower than the point-wise summation operation (d) while the feature split module of method (d) and (e) adopts our feature split operation (from 76.28 to 76.91, the difference is 0.63). This module not only has no effect on the feature integration part, but also inhibits the performance of the network.

In fact, two conclusions can be drawn from this phenomenon: 1) Without the operation of inter-channel information fusion, it is difficult for the network to adaptively split suitable features for the high- and low-frequency feature extraction. 2) Considering that the network can extract the high- and low-frequency information, the low-frequency information and high-frequency information are complementary to each other and usually do not appear in the same spatial position of the features. The point-wise summation is conducive to the integration of high- and low-frequency information. Besides the above description, we observe that the combination of the summation operation and our feature split module can acquire better performance, and the validation result of method (b) is significantly lower than that of method (d) (from 75.82 to 76.91). This suggests that the LFE and HFE branches are not completely suitable for the input information. The attention-based feature split module can strengthen the features that are suitable for extracting frequency information while weakening the inappropriate features.

\noindent \textbf{Down-sampling Methods}: We compare our strided-convolution method with the max-pooling down-sampling method, and can discover that the result of method (f) is better than that of method (d) (from 77.23 to 76.91). This means that the strided-convolution provides more complete frequency information than max-pooling. 

It is noted that we also visualize the spectrum maps and feature maps of the input data in Appendix. 

\section{Conclusion}
\label{conclusions}
In this work, we propose a dual-branch architecture, named FreConv, to extract the features of various frequencies, and integrate these features. A derivative-filter-like architecture architecture is designed to extract the information of high-frequency while a light extractor is employed. FreConv can exploit frequency information in a more reasonable way to enhance the representation ability of features and improve the network's efficiency. FreConv can replace vanilla convolution operation in-place, and it can save a substantial amount of parameters and FLOPS. Extensive experiments on various datasets, tasks, and network architectures demonstrated our FreConv's effectiveness. The self-supervised learning with network structure is currently a major focus in research society, we will apply FreConv to the self-supervised learning~\cite{li2021mst,li2022univip,li2023efficient}. Finally, we hope our FreConv can inspire research that exploits the information of whole frequencies.

\section{Acknowledgement}
This work was supported by National Natural Science Foundation of China under Grants 61976210, 62176254, 62006230, 62002357, 62206290 and National Key R$\&$D Program of China under Grant No.2021ZD0114600. We also thank for the valuable suggestions provided by Na Lu, Zhiyang Chen, and Kuan Zhu.

\bibliographystyle{IEEEbib}
\bibliography{icme2023template}

\end{document}